\documentclass[
twocolumn,
]{ceurart}

\usepackage{listings}

\usepackage[T1]{fontenc}
\usepackage[utf8]{inputenc}

\usepackage{multirow}
\usepackage{float}
\usepackage{graphicx}

\graphicspath{ {./images/} }

\author[1,2]{Krzysztof Olejniczak}[%
orcid=0000-0001-5390-5872,
]
\fnmark[1]

\address[1]{University of Oxford, United Kingdom}
\author[2]{Milan Šulc}[%
orcid=0000-0002-6321-0131,
]
\address[2]{Rossum.ai, Czech Republic}
\fntext[1]{The work was done when Krzysztof Olejniczak was an intern at Rossum.}
\begin{document}
\copyrightyear{2023}
\copyrightclause{Copyright for this paper by its authors.
  Use permitted under Creative Commons License Attribution 4.0
  International (CC BY 4.0).}

\conference{26th Computer Vision Winter Workshop, Robert Sablatnig and Florian Kleber (eds.), Krems, Lower Austria, Austria, Feb. 15-17, 2023}
\title{Text Detection Forgot About Document OCR}

\begin{abstract}
Detection and recognition of text from scans and other images, commonly denoted as \textit{Optical Character Recognition} (OCR), is a widely used form of automated document processing with a number of methods available. Yet OCR systems still do not achieve 100\% accuracy, requiring human corrections in applications where correct readout is essential. Advances in machine learning enabled even more challenging scenarios of text detection and recognition \textit{"in-the-wild"} -- such as detecting text on objects from photographs of complex scenes. While the state-of-the-art methods for \textit{in-the-wild} text recognition are typically evaluated on complex scenes, their performance in the domain of documents is typically not published, and a comprehensive comparison with methods for document OCR is missing. This paper compares several methods designed for \textit{in-the-wild} text recognition and for document text recognition, and provides their evaluation on the domain of structured documents. The results suggest that state-of-the-art methods originally proposed for \textit{in-the-wild} text detection also achieve competitive results on document text detection, outperforming available OCR methods. We argue that the application of document OCR should not be omitted in evaluation of text detection and recognition methods.
\end{abstract}

\begin{keywords}
  Text Detection \sep
  Text Recognition \sep
  OCR \sep
  Optical Character Recognition \sep
  Text In The Wild
\end{keywords}

\maketitle

\begin{figure*}[b!]
    \centering
    \includegraphics[width=1.00\textwidth]{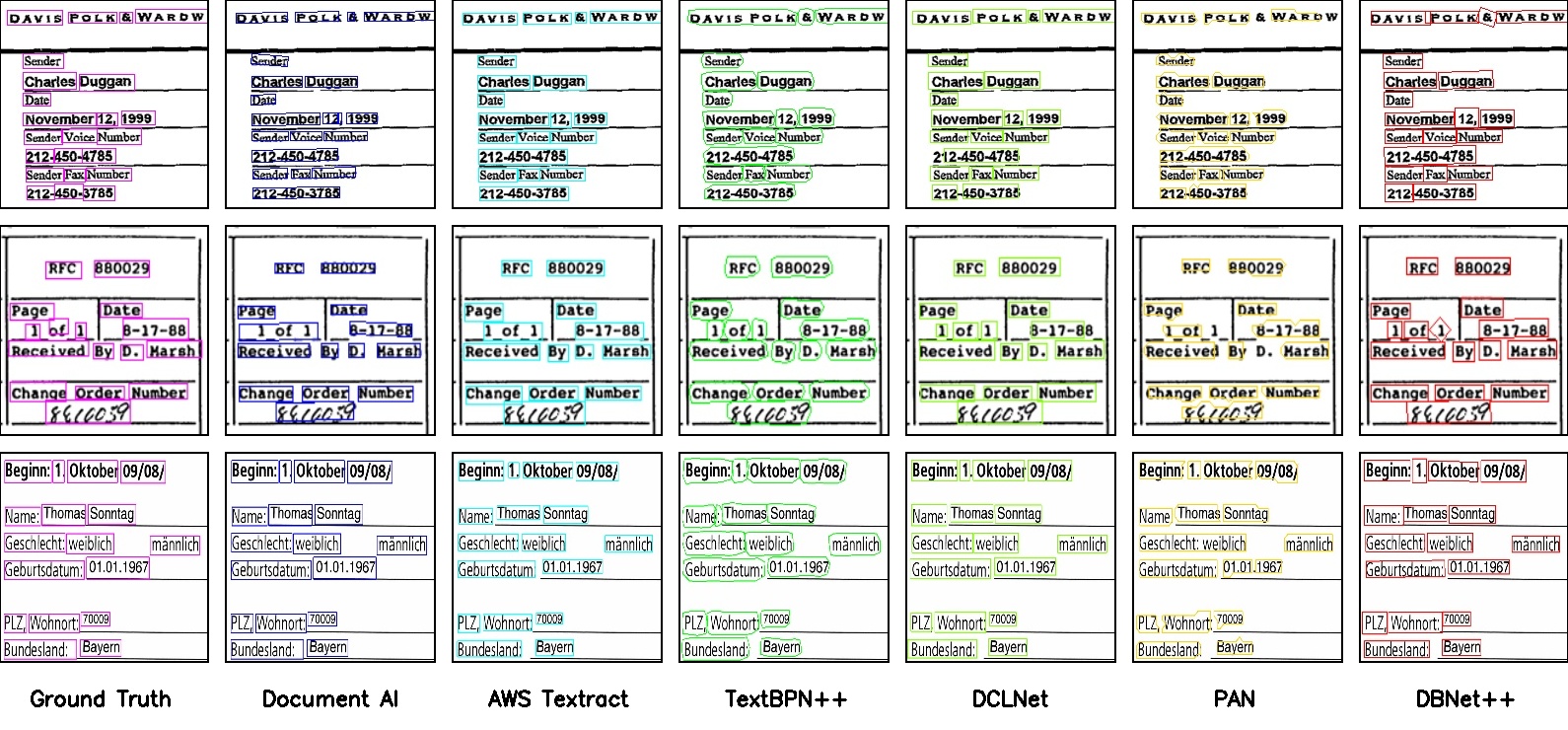}
    \caption{Comparison of detection results performed on documents from FUNSD dataset (cropped).}
    \label{fig:vis_comparison}
\end{figure*}

\section{Introduction}
\textit{Optical Character Recognition} (OCR) is a classic problem in machine learning and computer vision with standard methods \cite{PPOCR,tesseract} and surveys \cite{hamad2016detailed,hegghammer_2021,islam2017survey,memon2020handwritten} available. 
Recent advances in machine learning and its applications, such as autonomous driving, scene understanding or large-scale image retrieval, shifted the attention of Text Recognition research towards the more challenging \textit{in-the-wild} text scenarios, with arbitrarily shaped and oriented instances of text appearing in complex scenes. 
Spotting text \textit{in-the-wild} poses challenges such as extreme aspect ratios, curved or otherwise irregular text, complex backgrounds and clutter in the scenes. Recent methods \cite{DBNet++,TextBPN++} achieve impressive results on challenging text \textit{in-the-wild} datasets like TotalText \cite{totaltext} or CTW-1500 \cite{ctw1500}, with F1 reaching 90\% and 87\% respectively.
Although automated document processing remains one of the major applications of OCR, to the best of our knowledge, the results of \textit{in-the-wild} text detection models were never comprehensively evaluated on the domain of documents and compared with methods developed for document OCR. 
This paper reviews several recent Text Detection methods developed for the \textit{in-the-wild} scenario \cite{craft,dclnet,DBNet,DBNet++,PAN,TextBPN++}, evaluates their performance (out of the box and fine-tuned) on benchmark document datasets \cite{funsd,park2019cord,xu-etal-2022-xfund}, and compares their scores against popular Document OCR engines \cite{web_awstextract,web_documentai,tesseract}. Additionally, we adopt publicly available Text Recognition models \cite{sar,crnn} and combine them with Text Detectors to perform two-stage end-to-end text recognition for a complete evaluation of text extraction.

\section{Related Work}

\subsection{Document OCR}
OCR engines designed for the "standard" application domain of documents range from open-source projects such as TesseractOCR \cite{tesseract} and PP-OCR \cite{PPOCR} to commercial services, including AWS Textract \cite{web_awstextract} or Google Document AI \cite{web_documentai}. 
Despite Document OCR being a classic problem with many practical applications, studied for decades \cite{mori1999optical, herbert1982history}, it still cannot be considered 'solved' -- even the best engines struggle to achieve perfect accuracy.
The methodology behind the commercial cloud services is typically not disclosed. The most popular\footnote{Based on the GitHub repository \cite{web_tesseract_github} statistics.} open-source OCR engine at the time of publication, Tesseract \cite{tesseract} (v4 and v5), uses a \textit{Long Short-Term Memory} (LSTM) neural network as the default recognition engine.

\subsection{In-the-wild Text Detection}
\label{sec:detection}
\subsubsection{Regression-based Methods}
Regression-based Methods follow the object classification approach, reduced to a single-class problem.
TextBoxes \cite{textboxes} and TextBoxes++ \cite{textboxes++} locate text instances with various lengths by using sets of anchors with different aspect ratios.
Various regression-based methods utilize an iterative refinement strategy, iteratively enhancing the quality of detected boundaries.
LOMO \cite{lomo} uses an Iterative Refinement Module, which in every step regresses coordinates of each corner of the predicted boundary, with an attention mechanism.
PCR \cite{pcr} proposes a top-down approach, starting with predictions of centres and sizes of text instances, and iteratively improving the bounding boxes using its Contour Localisation Mechanism.
TextBPN++ \cite{TextBPN++} introduces an Iterative Boundary Deformation Module, utilizing Transformer Blocks with multi-head attention  \cite{transformer} encoder and a multi-layer perceptron decoder, to iteratively adjust vertices of detected instances.
Instead of considering vertices of the bounding boxes, DCLNet \cite{dclnet} predicts quadrilateral boundaries by locating four lines restricting the corresponding area, representing them in polar coordinates system.
To address the problem of arbitrary-shaped text detection and accurately model the boundaries of irregular text regions, more sophisticated bounding boxes representation ideas have been developed.
ABCNet \cite{abcnet} adapts cubic Bezier curves to parametrize curved text instances, gaining the possibility of fitting non-polygon shapes.
FCENet \cite{fcenet} proposes Fourier Contour Embedding method, predicting the Fourier signature vectors corresponding to the representation of the boundary in Fourier domain, and uses them to generate the shape of the instance with Inverse Fourier Transformation.

\subsubsection{Segmentation-based Methods}
Segmentation-based Methods aim to classify each pixel as either text or non-text, and generate bounding boxes using post-processing on so obtained pixel maps. The binary, deterministic nature of such pixel classification problem may cause learning confusion on the borders of text instances. Numerous methods address this issue by predicting text kernels (central regions of instances) and appropriately gathering pixels around them.
PSENet \cite{psenet} predicts kernels of different sizes and forms bounding boxes by iteratively expanding their regions.
PAN \cite{PAN} generates pixel classification and kernel maps, linking each classified text pixel to the nearest kernel.
CentripetalText \cite{sheng2021centripetaltext} produces centripetal shift vectors that map pixels to correct text centres.
KPN \cite{kpn} creates pixel embedding vectors, for each instance locates the central pixel and retrieves the whole shapes by measuring the similarities in embedding vectors with scalar product.
Vast majority of segmentation-based methods generate probability maps, representing how likely pixels are to be contained in some text region, and using certain binarization mechanism (e.g. by applying thresholding) convert them into binary pixel maps. However, the thresholds are often determined empirically, and uncareful choice of them may lead to drastic decrease in performance.
To solve this problem, DBNet \cite{DBNet} proposes a Differentiable Binarization Equation, making the step between probability and classification maps end-to-end trainable and therefore letting the network learn how to accurately binarise predictions.
DBNet++ \cite{DBNet++} further improves on the baseline by extending the backbone network with an Adaptive Scale Fusion attention module, enhancing the upscaling process and obtaining deeper features.
TextFuseNet \cite{textfusenet} generates features on three different levels: global-, word- and character-level, and fuses them to gain relevant context and deeper insight into the image structure.
Instead of detecting words, CRAFT \cite{craft} locates text on character-level, predicting the areas covered by single letters, and links characters of each instance with respect to the generated affinity map.

\section{Methods}
\subsection{Text Detection}
\label{sec:chosen_det_methods}
To cover a wide range of text detectors, we selected methods from Section \ref{sec:detection} with different approaches: 
for regression-based methods, we included TextBPN++ as a vertex-focused algorithm and DCLNet as an edge-focused approach. From segmentation-based methods, we selected DBNet and DBNet++ as pure segmentation and PAN as an approach linking text pixels to corresponding kernels.  
Finally, CRAFT was chosen as a character-level method.

\subsection{Text Recognition}
\label{sec:chosen_rec_methods}
The ultimate goal of text detection, especially in the case of document processing, is to recognize the text within the detected instances. Therefore, to evaluate the suitability of popular \textit{in-the-wild} detectors for document OCR, we perform end-to-end measurements with the following text recognition engines: SAR \cite{sar}, MASTER \cite{master} and CRNN \cite{crnn}. 
The open-source engines were combined with the detection methods in a two-stage manner: the input image was initially processed by a detector, which returned bounding boxes. Afterwards, the corresponding cropped instances were passed to recognition models. As a point of reference, we compare both the detection and end-to-end recognition results of the selected methods with predictions of three common engines for end-to-end document OCR: Tesseract \cite{tesseract}, Google Document AI \cite{web_documentai} and AWS Textract \cite{web_awstextract}.

\subsection{Metric}
To measure both detection and end-to-end performance, we used the CLEval \cite{cleval} metric.
Contrary to metrics such as \textit{Intersection over Union} (IoU) perceiving text on word-level, CLEval measures precision and recall on character level. As a consequence, it slightly reduces the punishment for splitting or merging problematic instances (e.g dates), providing reliable and intuitive comparison of the quality of detection and recognition. Additionally, the \textit{Recognition Score} evaluated by CLEval, approximately corresponding to the precision of character recognition, informs about the quality of the recognition engine specifically on the detected bounding boxes.

\begin{table*}[bth]
\caption{Document datasets used in the experiments for text detection and recognition.
}
\centering
\setlength{\tabcolsep}{3pt}
\begin{tabular}{|c|c|c|c|c|}
\hline
    \textbf{Dataset} & \textbf{Training images} & \textbf{Test images} & \textbf{Document Types} & \textbf{Language} \\

\hline \hline
    FUNSD \cite{funsd} & 149 & 50 & Distorted forms, surveys, reports & English \\
    CORD \cite{park2019cord} & 900 & 100 & Photos of Indonesian receipts & English   \\
     XFUND \cite{xu-etal-2022-xfund}  & 1043 & 350 & Clean scanned forms & Multilingual  \\
\hline 
\end{tabular} 
\label{table:datasets}
\end{table*}

\section{Experiments}
\subsection{Training Strategies}
DBNet \cite{DBNet}, DBNet++ \cite{DBNet++} and PAN \cite{PAN} were fine-tuned for 100 epochs (600 epochs in case of FUNSD)
with batch size of 8 and initial learning rate set to 0.0001 and decreasing by a factor of 10 at the 60th and 80th epoch (200th and 400th for FUNSD).
Baselines, pre-trained on SynthText \cite{synthtext} (DBNet, DBNet++) or ImageNet \cite{deng2009imagenet} (PAN), were downloaded from the MMOCR 0.6.2 Model Zoo \cite{web_mmocr_detection}.
\mbox{DCLNet \cite{dclnet}} was fine-tuned from a pre-trained model \cite{web_dclnet_github} on each dataset for
150 epochs with batch size of 4, initial learning rate of 0.001, decaying to 0.0001. 
For each dataset, TextBPN++ \cite{TextBPN++} was fine-tuned from a pre-trained model \cite{web_textbpn_github} for 50 epochs with batch size of 4, learning rate of 0.0001 and data augmentations consisting of flipping, cropping and rotations.
Given no publicly-available training scripts for CRAFT, during the experiments, we used the MLT model from the github repository \cite{web_craft_github} without fine-tuning. 
All experiments were performed using Adam optimizer with \mbox{momentum 0.9,} on a single GPU with 11 GB of VRAM (GeForce GTX-1080Ti).

\begin{table*}[t]
    \caption{Comparison of the detection performance of the chosen methods on benchmark datasets, with respect to the CLEval metric. "P", "R" and "F1" represent the precision, recall and \mbox{F1-score}, respectively.}
    \centering
    \begin{tabular}{|c|c|c|c|c|c|c|c|c|c|}
    \hline
        \multirow{3}{*}{\textbf{Method}}& 
        \multicolumn{3}{|c|}{\multirow{2}{*}{\textbf{FUNSD}}} &
        \multicolumn{3}{|c|}{\multirow{2}{*}{\textbf{CORD}}} &
        \multicolumn{3}{|c|}{\multirow{2}{*}{\textbf{XFUND}}}
        \\
         & \multicolumn{3}{|c|}{}& \multicolumn{3}{|c|}{}& \multicolumn{3}{|c|}{} \\
         \cline{2-10}
                                    &   P   &   R   &   F1   &   P   &   R   &   F1  &   P   &   R   &   F1  \\
        \hline \hline
        PAN \cite{PAN}              & 96.25 & 70.57 & 81.44 & 98.92&97.33&98.12 & 96.96 & 77.90 & 86.39 \\
        DBNet \cite{DBNet}          & 96.02 & 96.11 & 96.07 & 97.94 & 99.17 & 98.55 & 97.04 & 95.58 & 96.30 \\
        DBNet++ \cite{DBNet++}      & 97.45 & \textbf{97.40} & \textbf{97.42} & 97.58 & {99.60} & {98.58} & 97.87 & 97.93 & 97.90 \\
        TextBPN++ \cite{TextBPN++}  & 96.63 & 95.59 & 96.11 & 98.65 & \textbf{99.74} & \textbf{99.19} & 97.88 & 94.29 & 96.05 \\
        DCLNet \cite{dclnet}        & 94.16 & 95.35 & 94.75 & \textbf{98.67} & 97.91 & 98.29 & \textbf{98.22} & \textbf{98.17} & \textbf{98.20} \\
        CRAFT \cite{craft}          & \textbf{97.84} & 95.72 & 96.77 & 94.25 & 88.46 & 91.26 & 89.75 & 93.02 & 91.36  \\
        Tesseract \cite{tesseract}  & 80.13	& 73.80 & 76.83 & 76.46 & 47.38 & 58.51 & 85.84 & 87.47 & 86.65  \\
        Document AI \cite{web_documentai}  & 95.56 & 89.77 & 92.57 & 92.90 & 93.71& 93.30 & 89.49 & 90.68 & 90.08  \\
        AWS Textract \cite{web_awstextract} & 97.50 & 95.89 & 96.69 & 80.60 & 84.79 & 82.64 & 97.64& 88.14 & 92.65  \\
        \hline
    \end{tabular}
    \label{tab:detection_results}
\end{table*}
\subsection{Detection Results}
Results of the text detection methods selected in Section \ref{sec:chosen_det_methods} on the datasets from Table \ref{table:datasets} are presented in Table \ref{tab:detection_results}.
On FUNSD dataset, DBNet++ achieves both the highest detection recall (97.40\%) and \mbox{F1-score} (97.42\%).
The highest precision rate, 97.84\% was scored by CRAFT.
PAN performed the weakest out of all considered \textit{in-the-wild} algorithms, scoring just 81.44\% \mbox{F1-score}.
Despite having achieved better results on FUNSD, segmentation-based approaches were outperformed by regression-based methods on CORD and XFUND.
TextBPN++ proved to be the best performing algorithm on CORD in terms of recall and \mbox{F1-score}, scoring 99.74\% and 99.19\%, respectively.
DCLNet, for which the best precision rate on CORD (98.67\%) was recorded, achieved superior results on XFUND, outperforming the remaining methods with respect to all three measures: precision - 98.22\%, recall - 98.17\% and \mbox{F1-score} - 98.20\%.
Out of the considered popular engines for end-to-end document OCR, AWS Textract presented the best performance on the domain of scans of structured documents -- FUNSD and XFUND -- scoring 96.69\% and 92.65\% \mbox{F1-score}, respectively. Google Document AI generalized remarkably better to distorted photos of receipts from the CORD dataset, achieving 93.30\% F1-score, surpassing the scores of AWS Textract and Tesseract.
The results show that \textit{in-the-wild} detectors fine-tuned on document datasets can outperform popular OCR engines on the domain of structured documents in terms of the CLEval detection metric.
However, the results for the predictions of pre-trained detectors may not be fully representative due to differences in splitting rules. E.g. Document AI creates separate instances for special symbols, e.g. brackets, leading to undesired splitting of words like \textit{"name(s)"} into several fragments, lowering precision and recall. 
On all experimented datasets, all fine-tuned \textit{in-the-wild} text detection models reached high prediction scores, proving themselves capable of handling text in structured documents.
Qualitative analysis of detectors' predictions revealed that the major sources of error were incorrect splitting of long text fragments (e.g e-mail addresses), merging instances in dense text regions and missing short stand-alone text, such as single-digit numbers.

\begin{table*}[t]
    \caption{Comparison of the recognition performance of the chosen text detection methods combined \mbox{with MMOCR's} \cite{mmocr2021} SAR and MASTER default models, fine-tuned SAR, and docTR's \cite{doctr2021} CRNN default model, on FUNSD and CORD, with respect to the CLEval metric. "P", "R", "F1" and "S" represent the end-to-end precision, recall, \mbox{F1-score} and Recognition Score, respectively.}
    \centering
    \begin{tabular}{|c|c|c|c|c|c|c|c|c|c|}
    \hline
        \multirow{3}{*}{\textbf{Recognition}}&\multirow{3}{*}{\textbf{Detection}}& \multicolumn{4}{|c|}{\multirow{2}{*}{\textbf{FUNSD}}} & \multicolumn{4}{|c|}{\multirow{2}{*}{\textbf{CORD}}} \\
         & & \multicolumn{4}{|c|}{} & \multicolumn{4}{|c|}{}   \\ \cline{3-10}
                                   & &   P   &   R   &   F1   &   S &   P   &   R   &   F1   &   S\\
        
        \hline \hline
        
        \multirow{6}{*}{}
                        & PAN \cite{PAN}                & 76.14 & 74.17 & 75.14 & 79.79 & 82.04 & 84.27 & 83.14 & 84.76 \\
                        & DBNet \cite{DBNet}            & 79.10 & 82.51 & 80.77 & 83.33 & 82.76 & 85.79 & 84.25 & 85.49 \\
        SAR \cite{sar}  & CRAFT \cite{craft}            & 83.75 & 85.16 & 84.45 & 85.92 & 79.62 & 76.93 & 78.25 & 86.37 \\
        (baseline)        & TextBPN++ \cite{TextBPN++}    & 84.90 & 87.87 & 86.36 & 88.86 & 83.56 & 87.00 & 85.25 & 86.58 \\
                        & DBNet++ \cite{DBNet++}        & 80.04 & 83.53 & 81.75 & 82.85 & 82.95 & 86.66 & 84.76 & 85.89 \\
                        & DCLNet \cite{dclnet}          & 77.67 & 82.27 & 79.91 & 81.80 & 82.75 & 85.53 & 84.11 & 86.16 \\
        
        \hline
        
        \multirow{6}{*}{}
                        & PAN \cite{PAN}                & 86.37 & 76.61 & 81.20 & 90.23 & 87.73 & 88.95 & 88.34 & 90.59 \\
                        & DBNet \cite{DBNet}            & 87.48 & 88.07 & 87.77 & 91.90 & 91.12 & 94.00 & 92.54 & 94.02 \\
        SAR \cite{sar}  & CRAFT \cite{craft}            & 88.14 & 86.48 & 87.30 & 90.39 & 84.98 & 79.19 & 81.99 & 91.53 \\
        (fine-tuned)      & TextBPN++ \cite{TextBPN++}    & 88.12 & 88.32 & 88.22 & 92.16 & 91.46 & 96.21 & 93.77 & 94.77 \\
                        & DBNet++ \cite{DBNet++}        & 89.15 & 89.83 & 89.49 & 92.13 & 90.40 & 93.83 & 92.09 & 93.54 \\
                        & DCLNet \cite{dclnet}          & 86.10 & 87.30 & 86.70 & 90.46 & 87.69 & 90.02 & 88.84 & 91.58 \\
        
        \hline
        
        \multirow{6}{*}{{MASTER \cite{master}}}
        & PAN \cite{PAN}                & 77.50 & 74.58 & 76.01 & 81.10 & 90.25 & 92.12 & 91.17 & 93.16 \\
        & DBNet \cite{DBNet}            & 80.30 & 83.11 & 81.68 & 84.55 & 91.94 & 94.31 & 93.11 & 94.62 \\
        & CRAFT \cite{craft}            & 82.06 & 82.90 & 82.48 & 84.22 & 85.81 & 81.86 & 83.79 & 92.93 \\
        & TextBPN++ \cite{TextBPN++}    & 82.10 & 83.93 & 83.00 & 85.96 & 91.77 & 94.79 & 93.26 & 94.78 \\
        & DBNet++ \cite{DBNet++}        & 81.33 & 83.99 & 82.64 & 84.13 & 91.39 & 94.63 & 92.98 & 94.48 \\
        & DCLNet \cite{dclnet}          & 79.55 & 82.85 & 81.17 & 83.31 & 90.01 & 92.28 & 91.13 & 93.71 \\
        
        \hline
        
        \multirow{6}{*}{{CRNN \cite{crnn} }} & PAN \cite{PAN}& 90.31&87.14&88.70& 94.00 & 95.70 & 96.52 & 96.10 & 98.65 \\
        & DBNet \cite{DBNet}           & 89.07 & 91.56 & 90.30 & 93.24 & \textbf{96.00} & 97.51 & \textbf{96.75} & 98.67 \\
        & CRAFT \cite{craft}           & 91.20 & 91.67 & 91.43 & 93.40 & 93.12 & 87.25 & 90.09 & \textbf{98.73} \\
        & TextBPN++ \cite{TextBPN++}   & 89.94 & 91.80 & 90.86 & 93.86 & 95.35 & 97.71 & 96.52 & 98.48       \\
        & DBNet++ \cite{DBNet++}       & 90.97 & 93.52 & 92.23 & 93.71 & 95.43 & \textbf{97.85} & 96.62 & 98.51 \\
        & DCLNet \cite{dclnet}         & 89.84 & 92.95 & 91.37 & 93.16 & 95.04 & 96.34 & 95.69 & 98.52 \\
        \hline
        \multicolumn{2}{|c|}{Tesseract \cite{tesseract}} & 73.84 & 73.84 & 69.09 & 88.48 & 73.96 & 44.33 & 55.43 & 93.55 \\ \hline
        \multicolumn{2}{|c|}{Google Document AI \cite{web_documentai}} & 90.83 & 92.03 & 91.42 & 94.80 & 88.06 & 90.97 & 89.49 & 98.61 \\
        \hline
        \multicolumn{2}{|c|}{AWS Textract \cite{web_awstextract}}  & \textbf{93.61} & \textbf{95.46} & \textbf{94.53} & \textbf{95.78} & 84.53 & 82.13 & 83.32 & 96.63 \\
        \hline
    \end{tabular}
    \label{tab:recognition_results}
\end{table*}

\subsection{Recognition Results}
End-to-end text recognition results combining fine-tuned \textit{in-the-wild} detectors with SAR \cite{sar} and MASTER \cite{master} models from MMOCR 0.6.2 Model Zoo \cite{web_mmocr_recognition}, and CRNN \cite{crnn} from docTR \cite{doctr2021} are listed in Table \ref{tab:recognition_results}. 
The XFUND dataset was skipped for this experiment since it contains Chinese and Japanese characters, for which the recognition models were not trained. 
On FUNSD, the end-to-end measurement outcomes followed the patterns from detection: 
equipped with CRNN as the recognition engine, DBNet++ proved to be the best tuned model in terms of CLEval end-to-end Recall (93.52\%) and \mbox{F1-score} (92.23\%), losing only to CRAFT in terms of precision. Much higher \mbox{F1-score} (+2\%) was measured for AWS Textract, whose end-to-end results outperformed all of the considered algorithms. It is important to note that the Recognition Score for AWS Textract reached almost 96\%, surpassing CRNN's scores by c.a. 2\%. This suggests that the recognition engine used in AWS Textract, performing much more accurately on FUNSD than the CRNN model, may have been a crucial reason for the good results. When evaluated on CORD, models with Differentiable Binarization scored the highest marks in all end-to-end measures: recall (DBNet++), precision and \mbox{F1-score} (DBNet); significantly surpassing the remaining methods. Interestingly, despite obtaining the best recall rate, DBNet++ did not beat the simpler DBNet in terms of end-to-end \mbox{F1-score}. The predictions of regression-based approaches, better than segmentation-based ones when pure detection scores were measured, appeared to combine slightly worse with CRNN. TextBPN++, however, remained competitive, achieving similar results to DBNet and DBNet++. Recognition Scores of CRNN, regardless the choice of \textit{in-the-wild} detector, exceeded 93\% on FUNSD and 98.5\% on CORD, once again demonstrating the suitability of applying these algorithms to document text recognition. SAR model, not specifically trained on documents, presented poorer performance: the highest measured F1-scores on FUNSD and CORD were 86.36\% and 85.25\%, respectively, both obtained by the combination with TextBPN++. 
Fine-tuned SAR models achieved slightly higher F1-scores reaching 89.49\% on FUNSD (equipped with DBNet++ as the detector) and 93.77\% on CORD (combined with TextBPN++ detections). Despite gaining a noticeable advantage over the baseline, fine-tuned SAR models did not surpass the performance of the pre-trained CRNN. 
Similarly to SAR, the pre-trained MASTER model \cite{web_mmocr_recognition} worked the best in combination with TextBPN++, achieving F1 score of 83.00\% on FUNSD and 93.26\% on CORD.

\section{Conclusions}
Text detection research has witnessed great progress in recent years, thanks to advancements in deep machine learning. The recently introduced methods widened the range of possible applications of text detectors, making them viable for \textit{in-the-wild} text spotting. This shifted the attention towards more complex scenarios, including arbitrarily-shaped text or instances with non-orthogonal orientations. With automated document processing remaining one of the most relevant commercial OCR applications, we stress the importance of determining whether the state-of-the-art methods for scene text spotting can also improve document OCR. Our experiments prove that detectors designed for \textit{in-the-wild} text spotting can indeed be applied to documents with great success. In particular, fine-tuning models such as \mbox{DBNet++} or TextBPN++ yielded over 96\% detection \mbox{F1-score} on FUNSD, over 98\% detection \mbox{F1-score} on CORD and over 96\% detection \mbox{F1-score} on XFUND, with respect to the CLEval metric, outperforming Google Document AI and AWS Textract. Moreover, combining these detectors with a publicly-available CRNN recognition model in a two-stage manner consistently achieves over 90\% CLEval end-to-end \mbox{\mbox{F1-score}}, even without explicit fine-tuning of CRNN. We hope the results will bring more attention to evaluating future Text Detection methods not only in the \textit{text-in-the-wild} scenario, but also on the domain of documents.

\section*{Acknowledgement}
We acknowledge the help of Bohumír Zámečník, an expert on OCR systems, who helped with the supervision of Krzysztof's internship project.
\newpage
\bibliography{refs}

\begin{thebibliography}{46}
\expandafter\ifx\csname natexlab\endcsname\relax\def\natexlab#1{#1}\fi
\providecommand{\url}[1]{\texttt{#1}}
\providecommand{\href}[2]{#2}
\providecommand{\path}[1]{#1}
\providecommand{\DOIprefix}{doi:}
\providecommand{\ArXivprefix}{arXiv:}
\providecommand{\URLprefix}{URL: }
\providecommand{\Pubmedprefix}{pmid:}
\providecommand{\doi}[1]{\href{http://dx.doi.org/#1}{\path{#1}}}
\providecommand{\Pubmed}[1]{\href{pmid:#1}{\path{#1}}}
\providecommand{\bibinfo}[2]{#2}
\ifx\xfnm\relax \def\xfnm[#1]{\unskip,\space#1}\fi
%Type = Article
\bibitem[{Du et~al.(2020)Du, Li, Guo, Yin, Liu, Zhou, Bai, Yu, Yang, Dang, and
  Wang}]{PPOCR}
\bibinfo{author}{Y.~Du}, \bibinfo{author}{C.~Li}, \bibinfo{author}{R.~Guo},
  \bibinfo{author}{X.~Yin}, \bibinfo{author}{W.~Liu},
  \bibinfo{author}{J.~Zhou}, \bibinfo{author}{Y.~Bai}, \bibinfo{author}{Z.~Yu},
  \bibinfo{author}{Y.~Yang}, \bibinfo{author}{Q.~Dang},
  \bibinfo{author}{H.~Wang},
\newblock \bibinfo{title}{{PP-OCR:} {A} practical ultra lightweight {OCR}
  system},
\newblock \bibinfo{journal}{CoRR} \bibinfo{volume}{abs/2009.09941}
  (\bibinfo{year}{2020}). \URLprefix \url{https://arxiv.org/abs/2009.09941}.
  \href{http://arxiv.org/abs/2009.09941}{{\tt arXiv:2009.09941}}.
%Type = Article
\bibitem[{Kay(2007)}]{tesseract}
\bibinfo{author}{A.~Kay},
\newblock \bibinfo{title}{Tesseract: An open-source optical character
  recognition engine},
\newblock \bibinfo{journal}{Linux J.} \bibinfo{volume}{2007}
  (\bibinfo{year}{2007}) \bibinfo{pages}{2}.
%Type = Article
\bibitem[{Hamad and Mehmet(2016)}]{hamad2016detailed}
\bibinfo{author}{K.~Hamad}, \bibinfo{author}{K.~Mehmet},
\newblock \bibinfo{title}{A detailed analysis of optical character recognition
  technology},
\newblock \bibinfo{journal}{International Journal of Applied Mathematics
  Electronics and Computers}  (\bibinfo{year}{2016}) \bibinfo{pages}{244--249}.
%Type = Misc
\bibitem[{Hegghammer(2021)}]{hegghammer_2021}
\bibinfo{author}{T.~Hegghammer}, \bibinfo{title}{Ocr with tesseract, amazon
  textract, and google document ai: A benchmarking experiment},
  \bibinfo{year}{2021}. \URLprefix \url{osf.io/preprints/socarxiv/6zfvs}.
  \DOIprefix\doi{10.31235/osf.io/6zfvs}.
%Type = Article
\bibitem[{Islam et~al.(2017)Islam, Islam, and Noor}]{islam2017survey}
\bibinfo{author}{N.~Islam}, \bibinfo{author}{Z.~Islam},
  \bibinfo{author}{N.~Noor},
\newblock \bibinfo{title}{A survey on optical character recognition system},
\newblock \bibinfo{journal}{arXiv preprint arXiv:1710.05703}
  (\bibinfo{year}{2017}).
%Type = Article
\bibitem[{Memon et~al.(2020)Memon, Sami, Khan, and
  Uddin}]{memon2020handwritten}
\bibinfo{author}{J.~Memon}, \bibinfo{author}{M.~Sami}, \bibinfo{author}{R.~A.
  Khan}, \bibinfo{author}{M.~Uddin},
\newblock \bibinfo{title}{Handwritten optical character recognition (ocr): A
  comprehensive systematic literature review (slr)},
\newblock \bibinfo{journal}{IEEE Access} \bibinfo{volume}{8}
  (\bibinfo{year}{2020}) \bibinfo{pages}{142642--142668}.
%Type = Article
\bibitem[{Liao et~al.(2022)Liao, Zou, Wan, Yao, and Bai}]{DBNet++}
\bibinfo{author}{M.~Liao}, \bibinfo{author}{Z.~Zou}, \bibinfo{author}{Z.~Wan},
  \bibinfo{author}{C.~Yao}, \bibinfo{author}{X.~Bai},
\newblock \bibinfo{title}{Real-time scene text detection with differentiable
  binarization and adaptive scale fusion},
\newblock \bibinfo{journal}{IEEE Transactions on Pattern Analysis and Machine
  Intelligence}  (\bibinfo{year}{2022}).
%Type = Inproceedings
\bibitem[{Zhang et~al.(2021)Zhang, Zhu, Yang, Wang, and Yin}]{TextBPN++}
\bibinfo{author}{S.~Zhang}, \bibinfo{author}{X.~Zhu},
  \bibinfo{author}{C.~Yang}, \bibinfo{author}{H.~Wang},
  \bibinfo{author}{X.~Yin},
\newblock \bibinfo{title}{Adaptive boundary proposal network for arbitrary
  shape text detection},
\newblock in: \bibinfo{booktitle}{2021 {IEEE/CVF} International Conference on
  Computer Vision, {ICCV} 2021, Montreal, QC, Canada, October 10-17, 2021},
  \bibinfo{publisher}{{IEEE}}, \bibinfo{year}{2021}, pp.
  \bibinfo{pages}{1285--1294}.
%Type = Article
\bibitem[{Ch’ng et~al.(2020)Ch’ng, Chan, and Liu}]{totaltext}
\bibinfo{author}{C.~K. Ch’ng}, \bibinfo{author}{C.~S. Chan},
  \bibinfo{author}{C.~Liu},
\newblock \bibinfo{title}{Total-text: Towards orientation robustness in scene
  text detection},
\newblock \bibinfo{journal}{International Journal on Document Analysis and
  Recognition (IJDAR)} \bibinfo{volume}{23} (\bibinfo{year}{2020})
  \bibinfo{pages}{31--52}. \DOIprefix\doi{10.1007/s10032-019-00334-z}.
%Type = Article
\bibitem[{Liu et~al.(2019)Liu, Jin, Zhang, Luo, and Zhang}]{ctw1500}
\bibinfo{author}{Y.~Liu}, \bibinfo{author}{L.~Jin}, \bibinfo{author}{S.~Zhang},
  \bibinfo{author}{C.~Luo}, \bibinfo{author}{S.~Zhang},
\newblock \bibinfo{title}{Curved scene text detection via transverse and
  longitudinal sequence connection},
\newblock \bibinfo{journal}{Pattern Recognition} \bibinfo{volume}{90}
  (\bibinfo{year}{2019}) \bibinfo{pages}{337--345}. \URLprefix
  \url{https://www.sciencedirect.com/science/article/pii/S0031320319300664}.
  \DOIprefix\doi{https://doi.org/10.1016/j.patcog.2019.02.002}.
%Type = Inproceedings
\bibitem[{Baek et~al.(2019)Baek, Lee, Han, Yun, and Lee}]{craft}
\bibinfo{author}{Y.~Baek}, \bibinfo{author}{B.~Lee}, \bibinfo{author}{D.~Han},
  \bibinfo{author}{S.~Yun}, \bibinfo{author}{H.~Lee},
\newblock \bibinfo{title}{Character region awareness for text detection},
\newblock in: \bibinfo{booktitle}{Proceedings of the IEEE Conference on
  Computer Vision and Pattern Recognition}, \bibinfo{year}{2019}, pp.
  \bibinfo{pages}{9365--9374}.
%Type = Inproceedings
\bibitem[{Bi and Hu(2021)}]{dclnet}
\bibinfo{author}{Y.~Bi}, \bibinfo{author}{Z.~Hu},
\newblock \bibinfo{title}{Disentangled contour learning for quadrilateral text
  detection},
\newblock in: \bibinfo{booktitle}{Proceedings of the IEEE/CVF Winter Conference
  on Applications of Computer Vision}, \bibinfo{year}{2021}, pp.
  \bibinfo{pages}{909--918}.
%Type = Inproceedings
\bibitem[{Liao et~al.(2020)Liao, Wan, Yao, Chen, and Bai}]{DBNet}
\bibinfo{author}{M.~Liao}, \bibinfo{author}{Z.~Wan}, \bibinfo{author}{C.~Yao},
  \bibinfo{author}{K.~Chen}, \bibinfo{author}{X.~Bai},
\newblock \bibinfo{title}{Real-time scene text detection with differentiable
  binarization},
\newblock in: \bibinfo{booktitle}{Proc. AAAI}, \bibinfo{year}{2020}.
%Type = Article
\bibitem[{Wang et~al.(2019)Wang, Xie, Song, Zang, Wang, Lu, Yu, and Shen}]{PAN}
\bibinfo{author}{W.~Wang}, \bibinfo{author}{E.~Xie}, \bibinfo{author}{X.~Song},
  \bibinfo{author}{Y.~Zang}, \bibinfo{author}{W.~Wang},
  \bibinfo{author}{T.~Lu}, \bibinfo{author}{G.~Yu}, \bibinfo{author}{C.~Shen},
\newblock \bibinfo{title}{Efficient and accurate arbitrary-shaped text
  detection with pixel aggregation network},
\newblock \bibinfo{journal}{CoRR} \bibinfo{volume}{abs/1908.05900}
  (\bibinfo{year}{2019}). \URLprefix \url{http://arxiv.org/abs/1908.05900}.
  \href{http://arxiv.org/abs/1908.05900}{{\tt arXiv:1908.05900}}.
%Type = Inproceedings
\bibitem[{Guillaume~Jaume(2019)}]{funsd}
\bibinfo{author}{J.-P.~T. Guillaume~Jaume, Hazim Kemal~Ekenel},
\newblock \bibinfo{title}{Funsd: A dataset for form understanding in noisy
  scanned documents},
\newblock in: \bibinfo{booktitle}{Accepted to ICDAR-OST}, \bibinfo{year}{2019}.
%Type = Article
\bibitem[{Park et~al.(2019)Park, Shin, Lee, Lee, Surh, Seo, and
  Lee}]{park2019cord}
\bibinfo{author}{S.~Park}, \bibinfo{author}{S.~Shin}, \bibinfo{author}{B.~Lee},
  \bibinfo{author}{J.~Lee}, \bibinfo{author}{J.~Surh},
  \bibinfo{author}{M.~Seo}, \bibinfo{author}{H.~Lee},
\newblock \bibinfo{title}{Cord: A consolidated receipt dataset for post-ocr
  parsing}  (\bibinfo{year}{2019}).
%Type = Inproceedings
\bibitem[{Xu et~al.(2022)Xu, Lv, Cui, Wang, Lu, Florencio, Zhang, and
  Wei}]{xu-etal-2022-xfund}
\bibinfo{author}{Y.~Xu}, \bibinfo{author}{T.~Lv}, \bibinfo{author}{L.~Cui},
  \bibinfo{author}{G.~Wang}, \bibinfo{author}{Y.~Lu},
  \bibinfo{author}{D.~Florencio}, \bibinfo{author}{C.~Zhang},
  \bibinfo{author}{F.~Wei},
\newblock \bibinfo{title}{{XFUND}: A benchmark dataset for multilingual
  visually rich form understanding},
\newblock in: \bibinfo{booktitle}{Findings of the Association for Computational
  Linguistics: ACL 2022}, \bibinfo{publisher}{Association for Computational
  Linguistics}, \bibinfo{address}{Dublin, Ireland}, \bibinfo{year}{2022}, pp.
  \bibinfo{pages}{3214--3224}. \URLprefix
  \url{https://aclanthology.org/2022.findings-acl.253}.
  \DOIprefix\doi{10.18653/v1/2022.findings-acl.253}.
%Type = Misc
\bibitem[{Amazon(2022)}]{web_awstextract}
\bibinfo{author}{Amazon}, \bibinfo{title}{Amazon textract},
  \bibinfo{howpublished}{\url{https://aws.amazon.com/textract}},
  \bibinfo{year}{2022}. \bibinfo{note}{Accessed: 2022-09-25}.
%Type = Misc
\bibitem[{Google(2022)}]{web_documentai}
\bibinfo{author}{Google}, \bibinfo{title}{Google cloud document ai},
  \bibinfo{howpublished}{\url{https://cloud.google.com/document-ai}},
  \bibinfo{year}{2022}. \bibinfo{note}{Accessed: 2022-09-25}.
%Type = Article
\bibitem[{Li et~al.(2018)Li, Wang, Shen, and Zhang}]{sar}
\bibinfo{author}{H.~Li}, \bibinfo{author}{P.~Wang}, \bibinfo{author}{C.~Shen},
  \bibinfo{author}{G.~Zhang},
\newblock \bibinfo{title}{Show, attend and read: {A} simple and strong baseline
  for irregular text recognition},
\newblock \bibinfo{journal}{CoRR} \bibinfo{volume}{abs/1811.00751}
  (\bibinfo{year}{2018}). \URLprefix \url{http://arxiv.org/abs/1811.00751}.
  \href{http://arxiv.org/abs/1811.00751}{{\tt arXiv:1811.00751}}.
%Type = Article
\bibitem[{Shi et~al.(2015)Shi, Bai, and Yao}]{crnn}
\bibinfo{author}{B.~Shi}, \bibinfo{author}{X.~Bai}, \bibinfo{author}{C.~Yao},
\newblock \bibinfo{title}{An end-to-end trainable neural network for
  image-based sequence recognition and its application to scene text
  recognition},
\newblock \bibinfo{journal}{CoRR} \bibinfo{volume}{abs/1507.05717}
  (\bibinfo{year}{2015}). \URLprefix \url{http://arxiv.org/abs/1507.05717}.
  \href{http://arxiv.org/abs/1507.05717}{{\tt arXiv:1507.05717}}.
%Type = Book
\bibitem[{Mori et~al.(1999)Mori, Nishida, and Yamada}]{mori1999optical}
\bibinfo{author}{S.~Mori}, \bibinfo{author}{H.~Nishida},
  \bibinfo{author}{H.~Yamada}, \bibinfo{title}{Optical character recognition},
  \bibinfo{publisher}{John Wiley \& Sons, Inc.}, \bibinfo{year}{1999}.
%Type = Article
\bibitem[{Schantz(1982)}]{herbert1982history}
\bibinfo{author}{H.~F. Schantz},
\newblock \bibinfo{title}{The history of ocr, optical character recognition},
\newblock \bibinfo{journal}{Manchester Center, VT: Recognition Technologies
  Users Association}  (\bibinfo{year}{1982}).
%Type = Misc
\bibitem[{et~al.(2022)}]{web_tesseract_github}
\bibinfo{author}{S.~W. et~al.}, \bibinfo{title}{Tesseract open source ocr
  engine (main repository)},
  \bibinfo{howpublished}{\url{https://github.com/tesseract-ocr/tesseract}},
  \bibinfo{year}{2022}. \bibinfo{note}{Accessed: 2022-10-14}.
%Type = Inproceedings
\bibitem[{Liao et~al.(2017)Liao, Shi, Bai, Wang, and Liu}]{textboxes}
\bibinfo{author}{M.~Liao}, \bibinfo{author}{B.~Shi}, \bibinfo{author}{X.~Bai},
  \bibinfo{author}{X.~Wang}, \bibinfo{author}{W.~Liu},
\newblock \bibinfo{title}{Textboxes: {A} fast text detector with a single deep
  neural network},
\newblock in: \bibinfo{booktitle}{AAAI}, \bibinfo{year}{2017}.
%Type = Article
\bibitem[{Minghui~Liao and Bai(2018)}]{textboxes++}
\bibinfo{author}{B.~S. Minghui~Liao}, \bibinfo{author}{X.~Bai},
\newblock \bibinfo{title}{{TextBoxes++}: A single-shot oriented scene text
  detector},
\newblock \bibinfo{journal}{{IEEE} Transactions on Image Processing}
  \bibinfo{volume}{27} (\bibinfo{year}{2018}) \bibinfo{pages}{3676--3690}.
  \URLprefix \url{https://doi.org/10.1109/TIP.2018.2825107}.
  \DOIprefix\doi{10.1109/TIP.2018.2825107}.
%Type = Article
\bibitem[{Zhang et~al.(2019)Zhang, Liang, Huang, En, Han, Ding, and
  Ding}]{lomo}
\bibinfo{author}{C.~Zhang}, \bibinfo{author}{B.~Liang},
  \bibinfo{author}{Z.~Huang}, \bibinfo{author}{M.~En},
  \bibinfo{author}{J.~Han}, \bibinfo{author}{E.~Ding},
  \bibinfo{author}{X.~Ding},
\newblock \bibinfo{title}{Look more than once: An accurate detector for text of
  arbitrary shapes},
\newblock \bibinfo{journal}{CoRR} \bibinfo{volume}{abs/1904.06535}
  (\bibinfo{year}{2019}). \URLprefix \url{http://arxiv.org/abs/1904.06535}.
  \href{http://arxiv.org/abs/1904.06535}{{\tt arXiv:1904.06535}}.
%Type = Inproceedings
\bibitem[{Dai et~al.(2021)Dai, Zhang, Zhang, and Cao}]{pcr}
\bibinfo{author}{P.~Dai}, \bibinfo{author}{S.~Zhang},
  \bibinfo{author}{H.~Zhang}, \bibinfo{author}{X.~Cao},
\newblock \bibinfo{title}{Progressive contour regression for arbitrary-shape
  scene text detection},
\newblock in: \bibinfo{booktitle}{2021 IEEE/CVF Conference on Computer Vision
  and Pattern Recognition (CVPR)}, \bibinfo{year}{2021}, pp.
  \bibinfo{pages}{7389--7398}. \DOIprefix\doi{10.1109/CVPR46437.2021.00731}.
%Type = Article
\bibitem[{Vaswani et~al.(2017)Vaswani, Shazeer, Parmar, Uszkoreit, Jones,
  Gomez, Kaiser, and Polosukhin}]{transformer}
\bibinfo{author}{A.~Vaswani}, \bibinfo{author}{N.~Shazeer},
  \bibinfo{author}{N.~Parmar}, \bibinfo{author}{J.~Uszkoreit},
  \bibinfo{author}{L.~Jones}, \bibinfo{author}{A.~N. Gomez},
  \bibinfo{author}{L.~Kaiser}, \bibinfo{author}{I.~Polosukhin},
\newblock \bibinfo{title}{Attention is all you need},
\newblock \bibinfo{journal}{CoRR} \bibinfo{volume}{abs/1706.03762}
  (\bibinfo{year}{2017}). \URLprefix \url{http://arxiv.org/abs/1706.03762}.
  \href{http://arxiv.org/abs/1706.03762}{{\tt arXiv:1706.03762}}.
%Type = Article
\bibitem[{Liu et~al.(2020)Liu, Chen, Shen, He, Jin, and Wang}]{abcnet}
\bibinfo{author}{Y.~Liu}, \bibinfo{author}{H.~Chen}, \bibinfo{author}{C.~Shen},
  \bibinfo{author}{T.~He}, \bibinfo{author}{L.~Jin}, \bibinfo{author}{L.~Wang},
\newblock \bibinfo{title}{Abcnet: Real-time scene text spotting with adaptive
  bezier-curve network},
\newblock \bibinfo{journal}{CoRR} \bibinfo{volume}{abs/2002.10200}
  (\bibinfo{year}{2020}). \URLprefix \url{https://arxiv.org/abs/2002.10200}.
  \href{http://arxiv.org/abs/2002.10200}{{\tt arXiv:2002.10200}}.
%Type = Inproceedings
\bibitem[{Zhu et~al.(2021)Zhu, Chen, Liang, Kuang, Jin, and Zhang}]{fcenet}
\bibinfo{author}{Y.~Zhu}, \bibinfo{author}{J.~Chen},
  \bibinfo{author}{L.~Liang}, \bibinfo{author}{Z.~Kuang},
  \bibinfo{author}{L.~Jin}, \bibinfo{author}{W.~Zhang},
\newblock \bibinfo{title}{Fourier contour embedding for arbitrary-shaped text
  detection},
\newblock in: \bibinfo{booktitle}{CVPR}, \bibinfo{year}{2021}.
%Type = Inproceedings
\bibitem[{Wang et~al.(2019)Wang, Xie, Li, Hou, Lu, Yu, and Shao}]{psenet}
\bibinfo{author}{W.~Wang}, \bibinfo{author}{E.~Xie}, \bibinfo{author}{X.~Li},
  \bibinfo{author}{W.~Hou}, \bibinfo{author}{T.~Lu}, \bibinfo{author}{G.~Yu},
  \bibinfo{author}{S.~Shao},
\newblock \bibinfo{title}{Shape robust text detection with progressive scale
  expansion network},
\newblock in: \bibinfo{booktitle}{Proceedings of the IEEE Conference on
  Computer Vision and Pattern Recognition}, \bibinfo{year}{2019}, pp.
  \bibinfo{pages}{9336--9345}.
%Type = Inproceedings
\bibitem[{Sheng et~al.(2021)Sheng, Chen, and Lian}]{sheng2021centripetaltext}
\bibinfo{author}{T.~Sheng}, \bibinfo{author}{J.~Chen},
  \bibinfo{author}{Z.~Lian},
\newblock \bibinfo{title}{Centripetaltext: An efficient text instance
  representation for scene text detection},
\newblock in: \bibinfo{booktitle}{Thirty-Fifth Conference on Neural Information
  Processing Systems}, \bibinfo{year}{2021}.
%Type = Misc
\bibitem[{Zhang et~al.(2022)Zhang, Zhu, Hou, Yang, and Yin}]{kpn}
\bibinfo{author}{S.-X. Zhang}, \bibinfo{author}{X.~Zhu}, \bibinfo{author}{J.-B.
  Hou}, \bibinfo{author}{C.~Yang}, \bibinfo{author}{X.-C. Yin},
  \bibinfo{title}{Kernel proposal network for arbitrary shape text detection},
  \bibinfo{year}{2022}. \URLprefix \url{https://arxiv.org/abs/2203.06410}.
  \DOIprefix\doi{10.48550/ARXIV.2203.06410}.
%Type = Inproceedings
\bibitem[{Ye et~al.(2020)Ye, Chen, Liu, and Du}]{textfusenet}
\bibinfo{author}{J.~Ye}, \bibinfo{author}{Z.~Chen}, \bibinfo{author}{J.~Liu},
  \bibinfo{author}{B.~Du},
\newblock \bibinfo{title}{Textfusenet: Scene text detection with richer fused
  features},
\newblock in: \bibinfo{booktitle}{Proceedings of the Twenty-Ninth International
  Joint Conference on Artificial Intelligence, {IJCAI-20}},
  \bibinfo{publisher}{International Joint Conferences on Artificial
  Intelligence Organization}, \bibinfo{year}{2020}, pp.
  \bibinfo{pages}{516--522}.
%Type = Article
\bibitem[{Lu et~al.(2019)Lu, Yu, Qi, Chen, Gong, and Xiao}]{master}
\bibinfo{author}{N.~Lu}, \bibinfo{author}{W.~Yu}, \bibinfo{author}{X.~Qi},
  \bibinfo{author}{Y.~Chen}, \bibinfo{author}{P.~Gong},
  \bibinfo{author}{R.~Xiao},
\newblock \bibinfo{title}{{MASTER:} multi-aspect non-local network for scene
  text recognition},
\newblock \bibinfo{journal}{CoRR} \bibinfo{volume}{abs/1910.02562}
  (\bibinfo{year}{2019}). \URLprefix \url{http://arxiv.org/abs/1910.02562}.
  \href{http://arxiv.org/abs/1910.02562}{{\tt arXiv:1910.02562}}.
%Type = Article
\bibitem[{Baek et~al.(2020)Baek, Nam, Park, Lee, Shin, Baek, Lee, and
  Lee}]{cleval}
\bibinfo{author}{Y.~Baek}, \bibinfo{author}{D.~Nam}, \bibinfo{author}{S.~Park},
  \bibinfo{author}{J.~Lee}, \bibinfo{author}{S.~Shin},
  \bibinfo{author}{J.~Baek}, \bibinfo{author}{C.~Y. Lee},
  \bibinfo{author}{H.~Lee},
\newblock \bibinfo{title}{Cleval: Character-level evaluation for text detection
  and recognition tasks},
\newblock \bibinfo{journal}{CoRR} \bibinfo{volume}{abs/2006.06244}
  (\bibinfo{year}{2020}). \URLprefix \url{https://arxiv.org/abs/2006.06244}.
  \href{http://arxiv.org/abs/2006.06244}{{\tt arXiv:2006.06244}}.
%Type = Inproceedings
\bibitem[{Gupta et~al.(2016)Gupta, Vedaldi, and Zisserman}]{synthtext}
\bibinfo{author}{A.~Gupta}, \bibinfo{author}{A.~Vedaldi},
  \bibinfo{author}{A.~Zisserman},
\newblock \bibinfo{title}{Synthetic data for text localisation in natural
  images},
\newblock in: \bibinfo{booktitle}{IEEE Conference on Computer Vision and
  Pattern Recognition}, \bibinfo{year}{2016}.
%Type = Inproceedings
\bibitem[{Deng et~al.(2009)Deng, Dong, Socher, Li, Li, and
  Fei-Fei}]{deng2009imagenet}
\bibinfo{author}{J.~Deng}, \bibinfo{author}{W.~Dong},
  \bibinfo{author}{R.~Socher}, \bibinfo{author}{L.-J. Li},
  \bibinfo{author}{K.~Li}, \bibinfo{author}{L.~Fei-Fei},
\newblock \bibinfo{title}{Imagenet: A large-scale hierarchical image database},
\newblock in: \bibinfo{booktitle}{2009 IEEE conference on computer vision and
  pattern recognition}, \bibinfo{organization}{Ieee}, \bibinfo{year}{2009}, pp.
  \bibinfo{pages}{248--255}.
%Type = Misc
\bibitem[{Kuang et~al.(2022)Kuang, Sun, Li, Yue, Lin, Chen, Wei, Zhu, Gao,
  Zhang, Chen, Zhang, and Lin}]{web_mmocr_detection}
\bibinfo{author}{Z.~Kuang}, \bibinfo{author}{H.~Sun}, \bibinfo{author}{Z.~Li},
  \bibinfo{author}{X.~Yue}, \bibinfo{author}{T.~H. Lin},
  \bibinfo{author}{J.~Chen}, \bibinfo{author}{H.~Wei},
  \bibinfo{author}{Y.~Zhu}, \bibinfo{author}{T.~Gao},
  \bibinfo{author}{W.~Zhang}, \bibinfo{author}{K.~Chen},
  \bibinfo{author}{W.~Zhang}, \bibinfo{author}{D.~Lin}, \bibinfo{title}{Text
  detection models - mmocr 0.6.2 documentation},
  \bibinfo{howpublished}{\url{https://mmocr.readthedocs.io/en/latest/textdet_models.html}},
  \bibinfo{year}{2022}. \bibinfo{note}{Accessed: 2022-10-14}.
%Type = Misc
\bibitem[{Bi and Hu(2021)}]{web_dclnet_github}
\bibinfo{author}{Y.~Bi}, \bibinfo{author}{Z.~Hu}, \bibinfo{title}{Pytorch
  implementation of dclnet "disentangled contour learning for quadrilateral
  text detection"},
  \bibinfo{howpublished}{\url{https://github.com/SakuraRiven/DCLNet}},
  \bibinfo{year}{2021}. \bibinfo{note}{Accessed: 2022-10-13}.
%Type = Misc
\bibitem[{Zhang et~al.(2022)Zhang, Zhu, Yang, Wang, and
  Yin}]{web_textbpn_github}
\bibinfo{author}{S.~Zhang}, \bibinfo{author}{X.~Zhu},
  \bibinfo{author}{C.~Yang}, \bibinfo{author}{H.~Wang},
  \bibinfo{author}{X.~Yin}, \bibinfo{title}{Arbitrary shape text detection via
  boundary transformer},
  \bibinfo{howpublished}{\url{https://github.com/GXYM/TextBPN-Plus-Plus}},
  \bibinfo{year}{2022}. \bibinfo{note}{Accessed: 2022-09-29}.
%Type = Misc
\bibitem[{Baek et~al.(2019)Baek, Lee, Han, Yun, and Lee}]{web_craft_github}
\bibinfo{author}{Y.~Baek}, \bibinfo{author}{B.~Lee}, \bibinfo{author}{D.~Han},
  \bibinfo{author}{S.~Yun}, \bibinfo{author}{H.~Lee}, \bibinfo{title}{Official
  implementation of character region awareness for text detection (craft)},
  \bibinfo{howpublished}{\url{https://github.com/clovaai/CRAFT-pytorch}},
  \bibinfo{year}{2019}. \bibinfo{note}{Accessed: 2022-10-13}.
%Type = Article
\bibitem[{Kuang et~al.(2021)Kuang, Sun, Li, Yue, Lin, Chen, Wei, Zhu, Gao,
  Zhang, Chen, Zhang, and Lin}]{mmocr2021}
\bibinfo{author}{Z.~Kuang}, \bibinfo{author}{H.~Sun}, \bibinfo{author}{Z.~Li},
  \bibinfo{author}{X.~Yue}, \bibinfo{author}{T.~H. Lin},
  \bibinfo{author}{J.~Chen}, \bibinfo{author}{H.~Wei},
  \bibinfo{author}{Y.~Zhu}, \bibinfo{author}{T.~Gao},
  \bibinfo{author}{W.~Zhang}, \bibinfo{author}{K.~Chen},
  \bibinfo{author}{W.~Zhang}, \bibinfo{author}{D.~Lin},
\newblock \bibinfo{title}{Mmocr: A comprehensive toolbox for text detection,
  recognition and understanding},
\newblock \bibinfo{journal}{arXiv preprint arXiv:2108.06543}
  (\bibinfo{year}{2021}).
%Type = Misc
\bibitem[{Mindee(2021)}]{doctr2021}
\bibinfo{author}{Mindee}, \bibinfo{title}{doctr: Document text recognition},
  \bibinfo{howpublished}{\url{https://github.com/mindee/doctr}},
  \bibinfo{year}{2021}.
%Type = Misc
\bibitem[{Kuang et~al.(2021)Kuang, Sun, Li, Yue, Lin, Chen, Wei, Zhu, Gao,
  Zhang, Chen, Zhang, and Lin}]{web_mmocr_recognition}
\bibinfo{author}{Z.~Kuang}, \bibinfo{author}{H.~Sun}, \bibinfo{author}{Z.~Li},
  \bibinfo{author}{X.~Yue}, \bibinfo{author}{T.~H. Lin},
  \bibinfo{author}{J.~Chen}, \bibinfo{author}{H.~Wei},
  \bibinfo{author}{Y.~Zhu}, \bibinfo{author}{T.~Gao},
  \bibinfo{author}{W.~Zhang}, \bibinfo{author}{K.~Chen},
  \bibinfo{author}{W.~Zhang}, \bibinfo{author}{D.~Lin}, \bibinfo{title}{Text
  recognition models - mmocr 0.6.2 documentation},
  \bibinfo{howpublished}{\url{https://mmocr.readthedocs.io/en/latest/textrecog_models.html}},
  \bibinfo{year}{2021}. \bibinfo{note}{Accessed: 2022-10-14}.

\end{thebibliography}
\end{document}